\documentclass{article} 
\usepackage[T1]{fontenc}
\usepackage[final]{colm2026_conference}

\usepackage{microtype}
\usepackage{hyperref}
\usepackage{url}
\usepackage{booktabs}

\usepackage{hyperref}
\usepackage{listings}
\usepackage{rotating}
\usepackage{todonotes}
\usepackage{amsmath}
\usepackage{bbm}
\usepackage{caption}
\usepackage{subcaption}
\usepackage{wrapfig}
\usepackage{lineno}

\definecolor{darkblue}{rgb}{0, 0, 0.5}
\hypersetup{colorlinks=true, citecolor=darkblue, linkcolor=darkblue, urlcolor=darkblue}

\title{Evaluating the Retrieval Robustness of\\Large Language Models}

\author{Shuyang Cao, Karthik Radhakrishnan, David Rosenberg\\
\textbf{Steven Lu, Pengxiang Cheng}\\
Bloomberg\\
\texttt{\{scao152,kradhakris10,drosenberg44,slu126,pcheng134\}@bloomberg.net}\\
\AND
Lu Wang\\
University of Michigan\\
\texttt{wangluxy@umich.edu}\\
\And
Shiyue Zhang\thanks{Work done while at Bloomberg.}\\
Cohere\\
\texttt{shiyue.zhang@cohere.com}
}

\definecolor{codegreen}{rgb}{0,0.6,0}
\definecolor{codegray}{rgb}{0.5,0.5,0.5}
\definecolor{codepurple}{rgb}{0.58,0,0.82}
\definecolor{backcolour}{rgb}{0.95,0.95,0.92}

\lstdefinestyle{mystyle}{
    backgroundcolor=\color{backcolour},
    commentstyle=\color{codegreen},
    keywordstyle=\color{magenta},
    numberstyle=\tiny\color{codegray},
    stringstyle=\color{codepurple},
    basicstyle=\ttfamily\footnotesize,
    breakatwhitespace=false,
    breaklines=true,
    captionpos=b,
    keepspaces=true,
    numbers=left,
    numbersep=5pt,
    showspaces=false,
    showstringspaces=false,
    showtabs=false,
    tabsize=2
}

\newif\ifcomments
\ifcomments
\newcommand{\draftcomment}[3]{\ensuretext{\textcolor{#2}{[\ensuretext{\textcolor{#2}{\ensuremath{\textsc{#1}}}} #3]}}}
\else
\newcommand{\draftcomment}[3]{}
\fi

\begin{document}

\ifcolmsubmission
\linenumbers
\fi

\maketitle

\begin{abstract}

Retrieval-augmented generation (RAG) generally enhances large language models' (LLMs) ability to solve knowledge-intensive tasks. But RAG could also lead to performance degradation due to imperfect retrieval and the model's limited ability to leverage retrieved content.
In this work, we evaluate the robustness of LLMs in practical RAG setups (henceforth \emph{retrieval robustness}). We focus on three research questions: (1) whether RAG is always better than non-RAG; (2) whether more retrieved documents always lead to better performance; and (3) whether document order impacts results.
To facilitate this study, we establish a benchmark of 1,891 samples spanning five datasets across three task categories, each with documents retrieved using both sparse and dense retrievers. We introduce three robustness metrics, each corresponding to one research question.
Our experiments across 11 LLMs show that models achieve generally high retrieval robustness, but robustness varies substantially across tasks, suggesting that the decision to adopt RAG remains a case-by-case consideration.
We further examine four additional prompting strategies that vary how models interact with retrieved documents. We find that Qwen and GPT models suffer notable robustness declines when reasoning is disabled, even on single-hop QA tasks, and that providing retrieved documents as tool responses improves Claude models but hurts Qwen and GPT models, highlighting potential issues of the GPT models regardless of their best overall robustness under vanilla prompting.

\end{abstract}
\section{Introduction}

Large language models (LLMs) learn to acquire massive amounts of knowledge through large-scale pre-training, enabling them to answer knowledge-intensive questions~\citep{openai2024gpt4ocard, claude, llama33}.
However, relying exclusively on parametric knowledge can lead to inaccuracies when dealing with unseen or time-sensitive information, or when the model fails to precisely retrieve relevant knowledge from its own parameters.
To alleviate these limitations, retrieval-augmented generation (RAG) is proposed, where external documents containing information relevant to the task are fetched from a datastore and provided to the model as context during inference~\citep{chen-etal-2017-reading, RAG}.

Despite its potential, RAG does not always guarantee performance improvements.
The retriever might fail to retrieve relevant documents, and LLMs might be distracted by irrelevant content, leading to a performance drop~\citep{mallen-etal-2023-trust}.
As achieving perfect retrieval remains difficult in practice, it is crucial for LLMs to exhibit robust behavior in the RAG settings to mitigate risks in real-world deployment.

Previous work has shown that LLMs are particularly vulnerable when provided with noisy contexts that are synthetically constructed~\citep{chen2024benchmarking}. Distracted by the specially designed misleading information, models tend to produce incorrect outputs~\citep{wu2024how}.
While synthetic contexts provide useful insights, they differ from real-world retrieved contexts, which are typically sourced from credible corpora such as Wikipedia and trusted news outlets.

To bridge this gap, we benchmark LLMs' robustness under realistic RAG settings.
We consider an LLM \textit{retrieval-robust} if (1) its RAG performance matches or exceeds its non-RAG performance; (2) adding more retrieved documents leads to equal or better performance; and (3) its RAG performance is invariant to the order of retrieved documents.
Accordingly, we introduce three metrics: no-degradation rate, retrieval size robustness, and retrieval order robustness.

We curate a benchmark of 1{,}891 samples from five datasets spanning three task categories: open-domain QA (Natural Questions~\citep{kwiatkowski-etal-2019-natural}, HotpotQA~\citep{yang-etal-2018-hotpotqa}, MS MARCO~\citep{bajaj2018msmarco}), long-form QA (WixQA~\citep{wixqa}), and fact verification (SciFact-Open~\citep{wadden-etal-2022-scifact}).
\begin{wrapfigure}{r}{0.5\textwidth}
    \centering
    \includegraphics[width=0.48\textwidth]{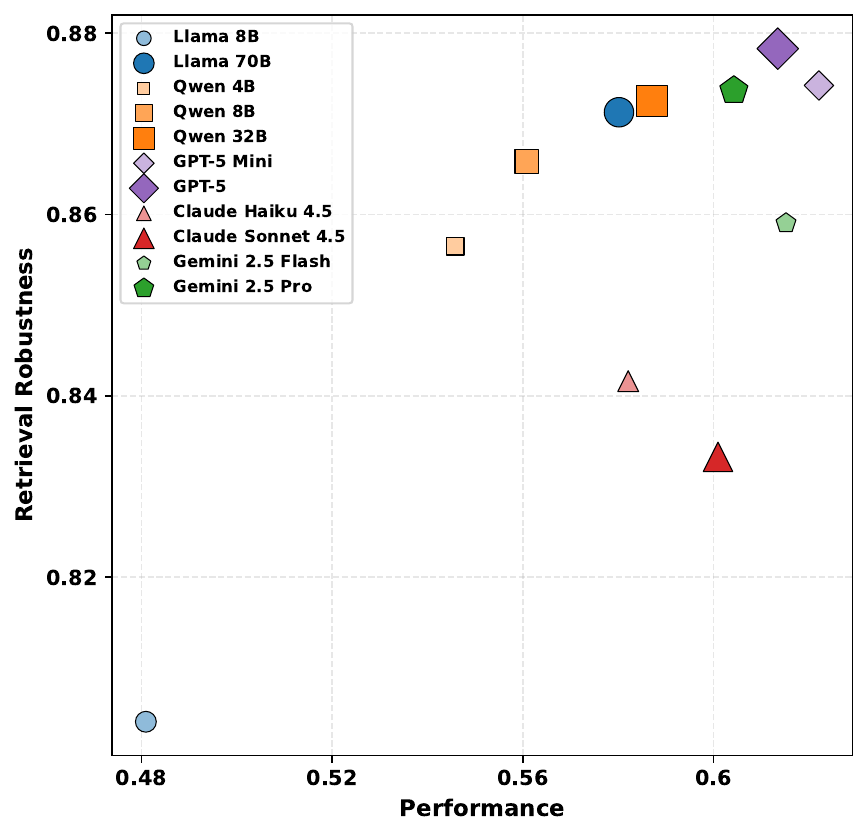}
    \caption{Comparison of retrieval robustness and task performance across various LLMs. The y-axis represents robustness (geometric mean of the three robustness metrics), while the x-axis represents task performance (average across all retrieval size $k$, retrieval order $o$, retrievers, and datasets).}
    \label{fig:geometric_mean_result}
\end{wrapfigure}
To construct retrieved contexts, we utilize two retrievers: a canonical sparse BM25~\citep{bm25} retriever and a dense retriever based on BGE~\citep{bge_embedding}.
For analyses of retrieval size and order robustness, RAG setups with multiple retrieval sizes (5 to 100 documents) and three ordering strategies (original, reversed, and reranked) are evaluated. Our experiments cover 11 LLMs from both open weight and proprietary families. Beyond vanilla prompting, we evaluate four additional strategies that vary how models interact with retrieved documents, including eliciting models' own knowledge, disabling reasoning, and structuring retrieval as tool calls.

We find that LLMs are generally robust, achieving over 75\% on the geometric mean of the three retrieval robustness metrics when prompting naively with retrieved contexts concatenated with the user queries, as shown in Figure~\ref{fig:geometric_mean_result}.
This suggests that, \emph{oftentimes}, (1) RAG outperforms non-RAG; (2) adding more retrieved documents improves performance; and (3) order of the documents has limited impact.
However, robustness varies substantially across tasks, with long-form generation showing notably lower robustness than short-form QA and fact verification, suggesting that the decision to adopt RAG should be task-specific.
Although GPT models achieve the best overall robustness,
our analysis with additional prompting strategies reveals that Qwen and GPT models suffer pronounced robustness drops when reasoning is disabled, even on simple single-hop QA tasks, suggesting their over-reliance on the reasoning context. 
We also find that providing retrieved documents as tool responses benefits Claude models but hurts Qwen and GPT, reflecting different optimization focuses across model families.

Our contributions can be summarized as follows:
\begin{itemize}
    \item We propose sample-level metrics to rigorously measure \textit{retrieval robustness}---how robustly LLMs handle queries in RAG setups.
    \item We compile a benchmark for evaluating retrieval robustness, following common RAG setups in practice. It comprises five datasets across three task categories, with documents retrieved using both sparse and dense retrievers.
    \item We conduct a comprehensive empirical study of 11 modern LLMs with five prompting strategies, demonstrating generally strong robustness in realistic settings while revealing that robustness is task-dependent and sensitive to model-specific design choices such as reasoning behaviors and tool-use optimization.
\end{itemize}
\section{Related Works}

\textbf{Retrieval Robustness.} Neural language models are shown to be easily distracted by adversarially inserted irrelevant content~\citep{jia-liang-2017-adversarial, 10.5555/3618408.3619699, weston20232attentionisneed}. However, irrelevant context comes in naturally in any RAG setup due to the imperfect retriever.  \citet{chen2024benchmarking} showed that the LLM-based RAG performance goes down when increasing the noise (i.e., documents that are relevant to the question but
do not contain any information about the answer) rate. \citet{wu2024how} conducted a deeper analysis and found that highly semantically related information is more likely to distract LLMs. \citet{thakur-etal-2024-knowing} evaluated LLM RAG performance with a completely irrelevant set of documents and observed non-trivial hallucination rates. 
On the contrary, \citet{cuconasu2024power} found that adding random documents to the prompt can actually improve LLM accuracy by up to 35\%, while irrelevant but high-scoring documents are more harmful, highlighting the nuanced relationship between retrieval quality and generation performance.  More recently, \citet{zeng2025worse} introduced RAGuard, a benchmark using real-world misinformation from Reddit, and found that all tested RAG systems perform \emph{worse} than their zero-shot baselines when exposed to misleading retrievals. In this work, we make the evaluation setup even more realistic and let the misleading or erroneous information occur naturally (no manual intervention) from imperfect retrieval and benchmark how robust modern LLMs are. 
Concretely, we follow the \textbf{\emph{retrieval robustness}} concept introduced by \citet{yoran2024making}, 
\begin{wrapfigure}[27]{r}{0.5\textwidth}
    \centering
    \includegraphics[width=0.44\textwidth]{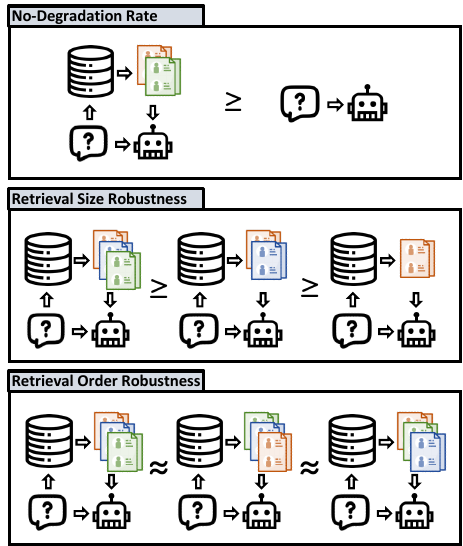}
    \caption{Our retrieval robustness metrics, targeting three research questions: (1) whether RAG is always better than non-RAG; (2) whether more retrieved documents always lead to better performance; (3) whether document orders lead to consistent results.}
    \label{fig:metrics}
\end{wrapfigure}
``retrieval-robust LLMs states that: 
(a) when relevant, the retrieved context should improve model performance; (b) when irrelevant, the retrieved context should not hurt model performance.'' 

\textbf{Improving RAG Robustness.} Many prior works aim to improve the robustness and reliability of RAG systems. Self-RAG~\citep{asai2024selfrag} trains a language model to adaptively decide when to retrieve, and to generate reflection tokens that critique its own outputs.
FLARE~\citep{jiang2023flare} takes a complementary approach by actively deciding \emph{when and what} to retrieve during generation, using low-confidence tokens as signals to trigger retrieval on-the-fly.
Corrective RAG (CRAG)~\citep{yan2024corrective} introduces a lightweight retrieval evaluator that assesses document quality and triggers corrective actions when retrieval confidence is low.
Chain-of-Note~\citep{yu2024chainofnote} trains models to generate sequential reading notes for each retrieved document, enabling systematic evaluation of document relevance before producing an answer.
InstructRAG~\citep{wei2024instructrag} instructs models to explicitly denoise retrieved documents through self-synthesized rationales.
Adaptive-RAG~\citep{jeong2024adaptiverag} dynamically selects among no retrieval, single-step retrieval, and multi-step retrieval strategies based on query complexity.
Astute RAG~\citep{wang2024astuterag} addresses the inevitability of imperfect retrieval by first eliciting the LLM's internal knowledge, then consolidating it with retrieved information while tracking source reliability.
\citet{fang2024raat} proposed adaptive adversarial training to improve noise robustness, and \citet{pan2024credibility} introduced credibility-aware generation that teaches LLMs to discern and weight information based on source reliability.
These methods are complementary to our work: while they propose solutions to improve robustness, we provide a systematic evaluation framework to measure it under realistic conditions. 

\section{Robustness Metrics}
\label{sec:metrics}

In this section, we present the three critical metrics for evaluating the retrieval robustness of an LLM system, illustrated in Figure~\ref{fig:metrics}.
We define an LLM system as a backbone LLM, paired with a prompting strategy.
Let $f(q, k, o)$ denote the performance of an LLM system, where $q$ is the task query, $k$ is the number of retrieved documents, and $o$ specifies the order of the retrieved documents.
In this paper, $f(q, k, o)$ is the correctness of the model's response to $q$, assessed by an LLM judge by comparing with the reference answer (\S\ref{sec:data_and_eval}).
When $k > 0$, $f(q, k, o)$ represents the performance of the LLM system in the RAG setup.
For consistency, we use $f(q, 0)$ to denote the performance of the LLM system in the non-RAG setup, where the model answers the query using its own knowledge.
See Section \ref{sec:rag_para} for the choices of $k$ and $o$ in our experiments.

\paragraph{No-Degradation Rate (NDR).}

This metric measures how often the LLM system's performance with RAG $f(q, k, o)$ (for any $k > 0$ and $o$) is at least as good as the performance without RAG $f(q, 0)$, which is calculated as:
\begin{equation}
    \text{NDR} = \frac{1}{Z} \sum_{\substack{q \in Q}} \sum_{\substack{k \in K}} \sum_{\substack{o \in O}} \mathbbm{1} \bigl[f(q, k, o) \;\ge\; f(q, 0)\bigr]
\end{equation}
where $K$ includes all choices of numbers of retrieved documents, $O$ represents all possible document orders used in the benchmark, and $Q$ is the set of all task samples. $Z = \lvert Q\rvert \cdot \lvert K\rvert \cdot \lvert O \rvert$ is the normalization factor for the aggregation.
A high NDR implies that, for most queries, using retrieval does not degrade performance relative to the non-RAG baseline. 

\paragraph{Retrieval Size Robustness (RSR).}

This metric examines how the system behaves as the number of retrieved documents increases.
Specifically, for each task query $q$ and each value of $k$, we check whether the performance is maintained or improved, compared to all smaller values of $k$.
RSR only considers $k > 0$, not involving the effect of NDR.
Results for various $k$s are then aggregated across all task samples, formally defined as:
\begin{equation}
\begin{aligned}
\text{RSR}_{(q, k_{i}, o)} &= \mathbbm{1} \bigl[\land_{j < i} [f(q, k_i, o) \;\ge\; f(q, k_{j}, o)]\bigr] \\
\text{RSR} &= \frac{1}{Z} \sum_{\substack{q \in Q}} \sum_{\substack{k_i \in K, i > 1}} \sum_{\substack{o \in O}} \text{RSR}_{(q, k_{i}, o)}
\end{aligned}
\end{equation}
where $Z = \lvert Q\rvert \cdot (\lvert K\rvert - 1) \cdot \lvert O \rvert$.
A high RSR indicates that performance rarely degrades when adding more retrieved documents.

\paragraph{Retrieval Order Robustness (ROR).}

ROR concerns the sensitivity of the system to the order of the same set of retrieved documents.
For a task sample $q$ and $k > 0$, let $O$ denote selected choices of permutations of the $k$ documents.
We can compute the standard deviation of the model performance over all permutations $o \in O$, which is represented as $\sigma_{o \in O} [f(q, k, o)]$.
For performance metrics bounded between 0 and 1, the standard deviation is bounded between 0 and 0.5. Therefore, we scale it by a factor of 2 to ensure the robustness metric ranges between 0 and 1.
We compute the ROR score as:
\begin{equation}
    \text{ROR} = \frac{1}{Z} \sum_{\substack{q \in Q}} \sum_{\substack{k \in K}} \bigl( 1 - 2\sigma_{o \in O} \bigl[f(q, k, o)\bigr] \bigr)
\end{equation}
where $Z = \lvert Q\rvert \cdot \lvert K\rvert$. A higher ROR means that different permutations of the same set of documents produce more consistent performance.

The three metrics capture complementary aspects of retrieval robustness, reflecting different desired behaviors of LLM systems with RAG in real world applications.
NDR provides a safety guarantee that retrieval will not harm results;
RSR is critical for scenarios where retrieval size can be scaled up for enhanced performance;
and ROR is important for situations where document ranking is imperfect.
Note that, for simplicity, we omit the marginalization over two different retrievers (see Section~\ref{sec:rag_para}) from the equations of all three metrics. 

\section{Benchmark Setups}
\label{sec:benchmark}

\subsection{Data and Evaluation Metrics}
\label{sec:data_and_eval}

\paragraph{Tasks.} We sample from five datasets spanning three task categories.
For \textit{open-domain QA}, we use Natural Questions~\citep{kwiatkowski-etal-2019-natural}, which contains samples derived from Google Search queries, covering a broad range of questions real-world users ask online; HotpotQA~\citep{yang-etal-2018-hotpotqa}, a multi-hop QA dataset that requires chaining multiple passages to answer questions; and MS MARCO~\citep{bajaj2018msmarco}, a large-scale reading comprehension dataset derived from real Bing search queries.
For \textit{long-form QA}, we use WixQA~\citep{wixqa}, an enterprise QA benchmark grounded in real customer support interactions from the Wix Help Center.
For \textit{fact verification}, we use SciFact-Open~\citep{wadden-etal-2022-scifact}, which targets open-domain scientific claim verification against a large corpus of research abstracts.
We randomly sample 500 examples each from Natural Questions, HotpotQA, and MS MARCO. For WixQA and SciFact-Open, we use all available samples: 200 and 191, respectively, totaling 1{,}891 samples.

\paragraph{Evaluation Metrics.}
Previous work usually used string match metrics for answer evaluation~\citep{mallen-etal-2023-trust, gao-etal-2023-enabling}. However, such metrics are rigid and cannot reliably evaluate model performance.
For \textit{open-domain QA} tasks (Natural Questions, HotpotQA, MS MARCO), we combine traditional string matching with LLM judges. We first check whether the gold answer can be found as a substring in the model-generated output. For samples where the gold answer is not found, we use three LLM judges---GPT-5-mini~\citep{gpt5}, Gemini-2.5-Flash~\citep{gemini25}, and Claude Haiku~4.5~\citep{claude45}---and take the majority vote to determine whether the generated answer semantically aligns with the gold answer (see the prompts in Appendix~\ref{app:prompt}).
For \textit{long-form QA} (WixQA), we follow the original authors and use ROUGE-2 recall~\citep{lin-2004-rouge} as the evaluation metric~\citep{wixqa}.
The \textit{fact verification} task (SciFact-Open) only requires a binary judgment of whether the claim is supported by the retrieved documents, and thus we use the exact match of the model's binary judgment with the gold label.

While using LLM judges for all tasks would be better (e.g., evaluating the coverage of reference facts for long-form QA), the large number of experimental setups (models $\times$ retrieval sizes $\times$ orders $\times$ retrievers $\times$ datasets) makes purely LLM-based evaluation prohibitively costly.
As a supplementary analysis, in Appendix~\ref{app:wixqa_llm_judge}, on a subset of WixQA, we report evaluation results that are based on LLM judges.
In addition, we include an analysis of the agreement level between LLM judges in Appendix~\ref{app:judge_agreement}.

\paragraph{Retrieval Corpus.}
For Natural Questions and HotpotQA, we use Wikipedia as the retrieval corpus. We processed the wikidump from November 2023, which contains 6 million articles. We split each article into chunks by double newlines, resulting in 20 million chunks. Each chunk is treated as an independent "document" for retrieval.
For MS MARCO, WixQA, and SciFact-Open, we use the passage corpora provided with each respective dataset.

\subsection{LLM Systems}

\paragraph{Backbone LLMs.}

11 LLMs from two open weight families and three proprietary families are tested, including Qwen3 (4B, 8B, 32B)~\citep{qwen3}, Llama-3 Instruct (3.1-8B, 3.3-70B)~\citep{llama31, llama33}, OpenAI GPT-5 and GPT-5-mini~\citep{gpt5}, Google Gemini-2.5-Pro and Gemini-2.5-Flash~\citep{gemini25}, and Anthropic Claude Sonnet~4.5 and Claude Haiku~4.5~\citep{claude45}.
After extending the context length of Qwen3 models to 128K with RoPE scaling following official recommendations, all models are able to consume the maximum retrieval size of 100 documents.

\paragraph{Prompting Strategies.}

Besides the vanilla prompting strategy that concatenates all retrieved documents in the user prompt, we explore four alternative strategies that vary how models interact with the retrieved documents:

\textbf{OwnKnow} obtains a draft answer based on models' own knowledge by prompting without retrieval in the first step, and then inserts this draft answer into the prompt for the RAG setup. This resembles how Astute RAG~\citep{wang2024astuterag} elicits the LLM's internal knowledge.

Most of the latest LLMs have reasoning capability. \textbf{NoThink} disables or minimizes the model's reasoning budget (i.e., sets the thinking token budget to the lowest possible), testing how disabling reasoning impacts models' retrieval robustness profiles.

As using LLMs as tool-use agents has become one of the most prevalent usage patterns, we also examine how structuring retrieval results as tool responses affects the robustness profile.
\textbf{ForcedTool} creates a dummy retrieval tool and forces the model to call it, and the retrieved documents are then provided as tool responses rather than directly in the user prompt.
\textbf{AutoTool} is similar to ForcedTool but lets the model decide whether to call the retrieval tool, which can be considered as an agent variant of the OwnKnow strategy.

As Llama models do not natively support tool usage and adjustment of reasoning content, we only test Llama models with vanilla prompting and OwnKnow.

\subsection{RAG Parameters}
\label{sec:rag_para}

\begin{wrapfigure}[20]{r}{0.5\textwidth}
    \centering
    \includegraphics[width=0.48\textwidth]{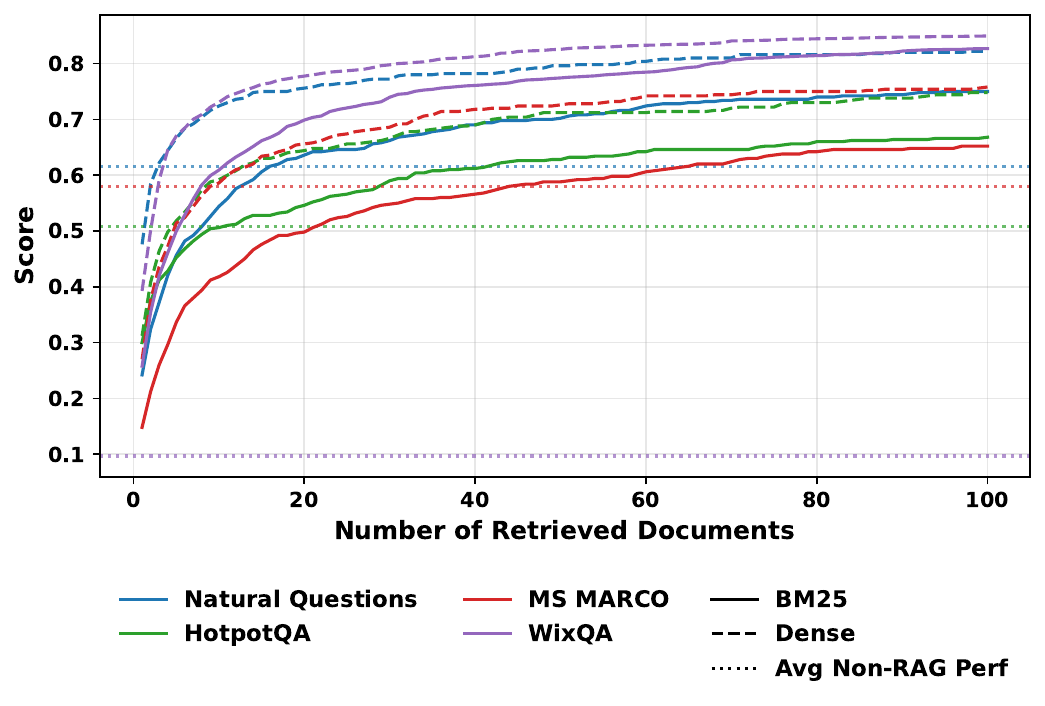}
    \caption{Performance of the retrievers, measured by the recall of gold answers within the concatenated retrieved documents. For open-domain QA, the gold answer is considered covered if any of its alternative forms exactly match a substring in the concatenated retrieved documents. For WixQA, we calculate the ROUGE-2 recall.}
    \label{fig:rag_upperbound}
    \vspace{-7pt}
\end{wrapfigure}
\paragraph{Retrievers.} Our retrieval system is built on top of Apache Solr 9\footnote{\url{https://solr.apache.org/docs/9 _0_0/index.html}}. We use two retrievers: one is the canonical sparse retriever based on BM25~\citep{bm25}, and the other is a cosine similarity based dense retriever where we embedded each document by bge-large-en-v1.5\footnote{\url{http://huggingface.co/BAAI/bge-large-en-v1.5}}~\citep{bge_embedding}. For any robustness metric defined in Section~\ref{sec:metrics}, we get the results for both retrievers and take the average.

\paragraph{Sizes.}

We experiment with retrieval sizes of 5, 10, 25, 50, 75, and 100 documents.
The retrieval size is capped at 100 documents as most models have reached their maximum context lengths. When the retrieved documents exceed the maximum context length of a model, we iteratively drop the lowest ranked document.

\paragraph{Orders.} For each of these sizes, we apply three ordering strategies based on the retriever's ranking of the documents: the \textbf{original} order (returned by the retriever), the \textbf{reversed} order (reversing the original order), and a \textbf{reranked} order produced by the BGE reranker~\citep{chen2024bge}.\footnote{\url{https://huggingface.co/BAAI/bge-reranker-v2-m3}}
We test the reversed order because sometimes we want to put the most relevant document at the end of the prompt (the closest to the answer). We include the reranked order to simulate a realistic reranking stage on top of the retriever.

\paragraph{Retrieval Quality.}

As our retrieval robustness benchmark relies on the retrievers, we examine the retrieval quality by checking the recall of gold answers within the retrieved documents.
We follow prior work and determine that the concatenated retrieved documents contain the gold answer if its substring is an exact match of any form of the gold answer (substring exact match)~\citep{mallen-etal-2023-trust}.
For WixQA, we calculate the ROUGE-2 recall by taking the concatenated retrieved documents as the hypotheses.
SciFact is excluded from this retrieval quality analysis as it only contains gold binary labels.
For reference, we also report the average model performance without RAG  (Non-RAG Perf).
As shown in Figure~\ref{fig:rag_upperbound}, both retrievers provide sufficiently high-quality retrieval, ensuring that the findings of our experiments are based on valid setups.

\begin{figure*}[t]
    \centering
    \includegraphics[width=0.98\textwidth]{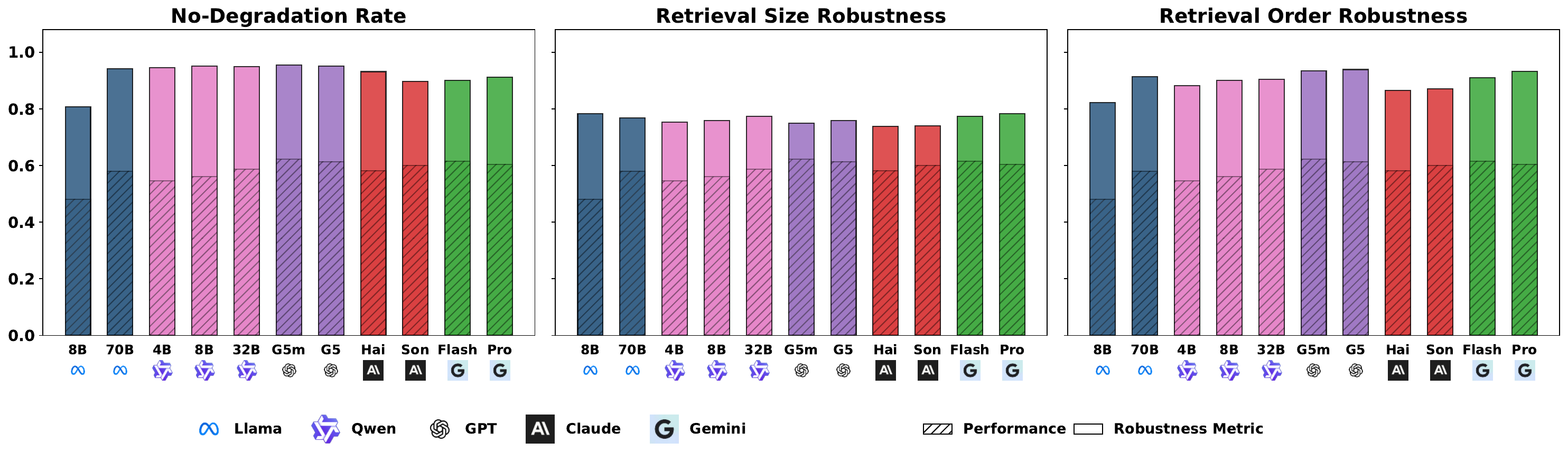}
    \caption{The three retrieval robustness metrics and task performance of experimented LLMs using vanilla prompting. G5m: GPT-5 Mini; G5: GPT-5; Hai: Haiku; Son: Sonnet. The mean of task performance achieved with different retrieval sizes and orders is shown for each model. Model families are indicated by icons, while the variants are indicated by model sizes or names. Models generally have good retrieval robustness. 
    }
    \label{fig:robustness_separate_result}
\end{figure*}

\section{Results}

\subsection{Overall Robustness}

We report the three retrieval robustness metrics for LLM systems using vanilla prompting in Figure~\ref{fig:robustness_separate_result}.
Besides robustness, task performance is shown in the same figure with bars with a different hatch style. Retrieval robustness is calculated following the definitions in Section~\ref{sec:metrics}, while task performance is the average score across all retrieval size $k$, retrieval order $o$, retrievers, and datasets.
\textbf{All models achieve higher than 75\% retrieval robustness across all metrics.}
Compared to prior studies that highlight the weak robustness of RAG systems under synthetic setups, such as using artificially created documents~\citep{wu2024how}, we show that LLMs demonstrate surprisingly good retrieval robustness in more realistic settings.

\begin{figure*}[t]
    \centering
    \includegraphics[width=0.98\textwidth]{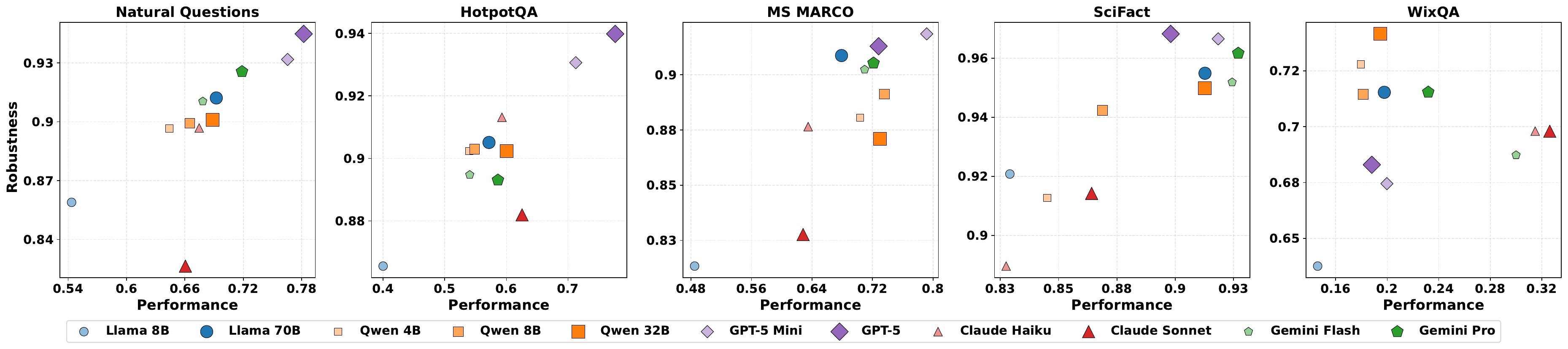}
    \caption{Geometric mean of the three retrieval robustness metrics compared against task performance for each model using vanilla prompting across various datasets.
    On WixQA, models exhibit low retrieval robustness because the task requires generating open-ended long-form outputs. These outputs are inherently high-variance and are further affected by the long input context constructed with the retrieved documents.}
    \label{fig:dataset_scatter_result}
\end{figure*}

Nevertheless, \textbf{retrieval robustness varies substantially across datasets} (Figure~\ref{fig:dataset_scatter_result}), suggesting that the decision to adopt RAG remains a case-by-case consideration. 
While models generally demonstrate high retrieval robustness on short-form question answering and fact verification, their robustness drops significantly on WixQA, which requires longer outputs.
This is partly because models already show high variance when generating open-ended, long-form content, and the inclusion of many retrieved documents can further increase this variance.
Furthermore, even on tasks where models are generally robust, the top-performing model is not consistent. For example, on short-form datasets, GPT-5 typically shows the best robustness. 
However, on MS MARCO, its smaller and more cost-effective variant actually surpasses it.

\subsection{Effects of Prompting}

\begin{figure*}[t]
    \centering
    \includegraphics[width=0.80\textwidth]{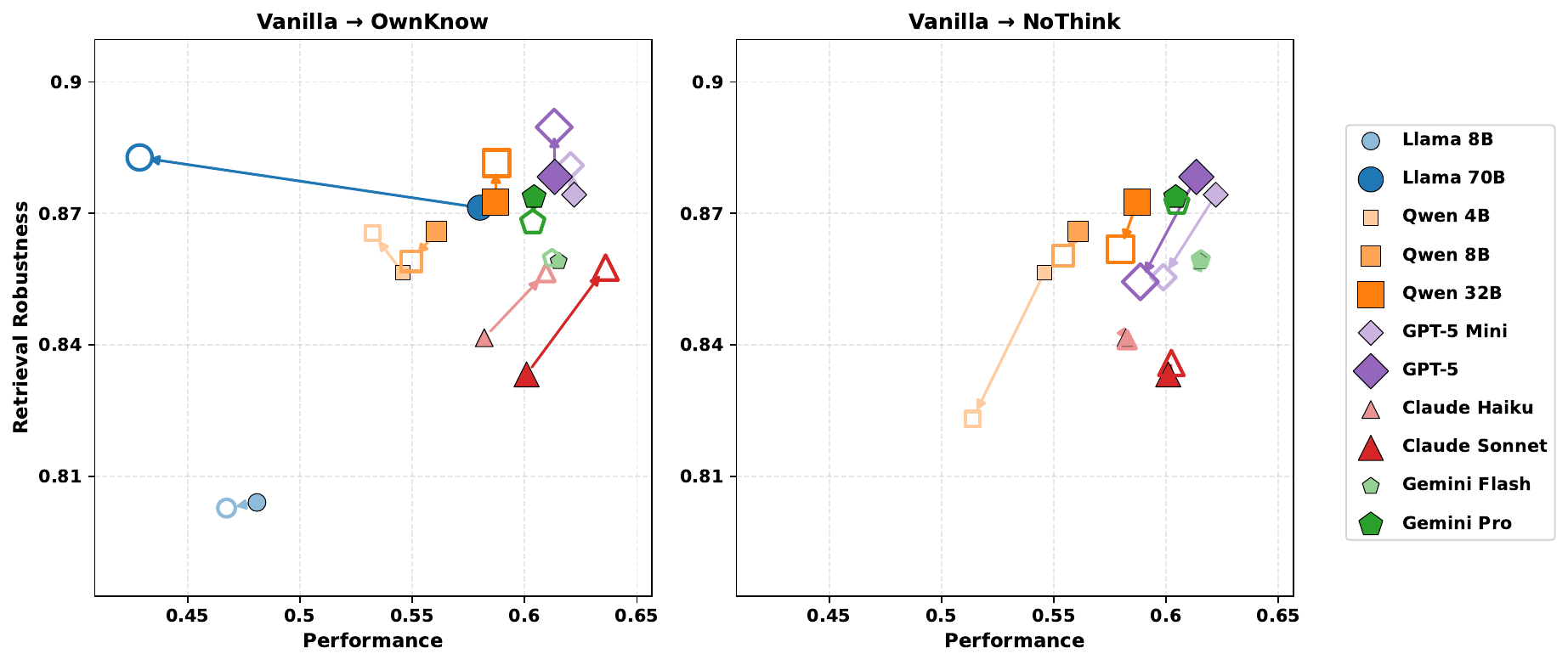}
    \caption{Effects of OwnKnow and NoThink settings. Shaded markers represent vanilla prompting, while transparent markers represent OwnKnow / NoThink.
    Effects are indicated by arrows pointing from the vanilla setting to OwnKnow / NoThink setting. Generally speaking, providing models with their prior knowledge and enabling reasoning improves robustness.}
    \label{fig:effect_ownknow_nothink}
\end{figure*}

Figure~\ref{fig:effect_ownknow_nothink} visualizes how adding models' prior knowledge to the context (\textbf{OwnKnow}) or disabling reasoning (\textbf{NoThink}) impacts model performance and retrieval robustness.
OwnKnow generally provides modest robustness gains, as the non-RAG draft serves as an anchor that reduces variance in model outputs.
This is analogous to self-refinement, which has been shown to be an effective prompting technique~\citep{yang-etal-2022-re3,madaan2023selfrefineiterativerefinementselffeedback}.
However, the inclusion of models' prior knowledge naturally reduces the tendency of models to use information in the retrieved contexts, which can lead to drops of performance compared to the vanilla prompting.
Besides the huge performance drop of Llama-3.3-70B as shown in the figure, we look into the maximum task performance across retrieval sizes rather than the mean task performance, and observe that all models except for the Claude family yield a lower maximum task performance when using OwnKnow.

Among models explicitly trained or fine-tuned for reasoning, Qwen and GPT experience more pronounced declines in both robustness and performance than other model families when reasoning mode is disabled.
A detailed breakdown across various datasets (Figure~\ref{fig:effect_nothink_per_dataset} in Appendix~\ref{app:effect_per_dataset}) reveals that while both Qwen and GPT experience the expected substantial performance and robustness drops on HotpotQA, a multi-hop QA dataset that requires some reasoning, they also suffer substantial robustness declines on simpler tasks.
Most concerning is their robustness on single-hop datasets, including Natural Questions and MS MARCO, which suggests these models may have become overly dependent on their reasoning context even for basic RAG tasks.

\subsection{Effects of Tool Usage}

\begin{figure*}[t]
    \centering
    \includegraphics[width=0.80\textwidth]{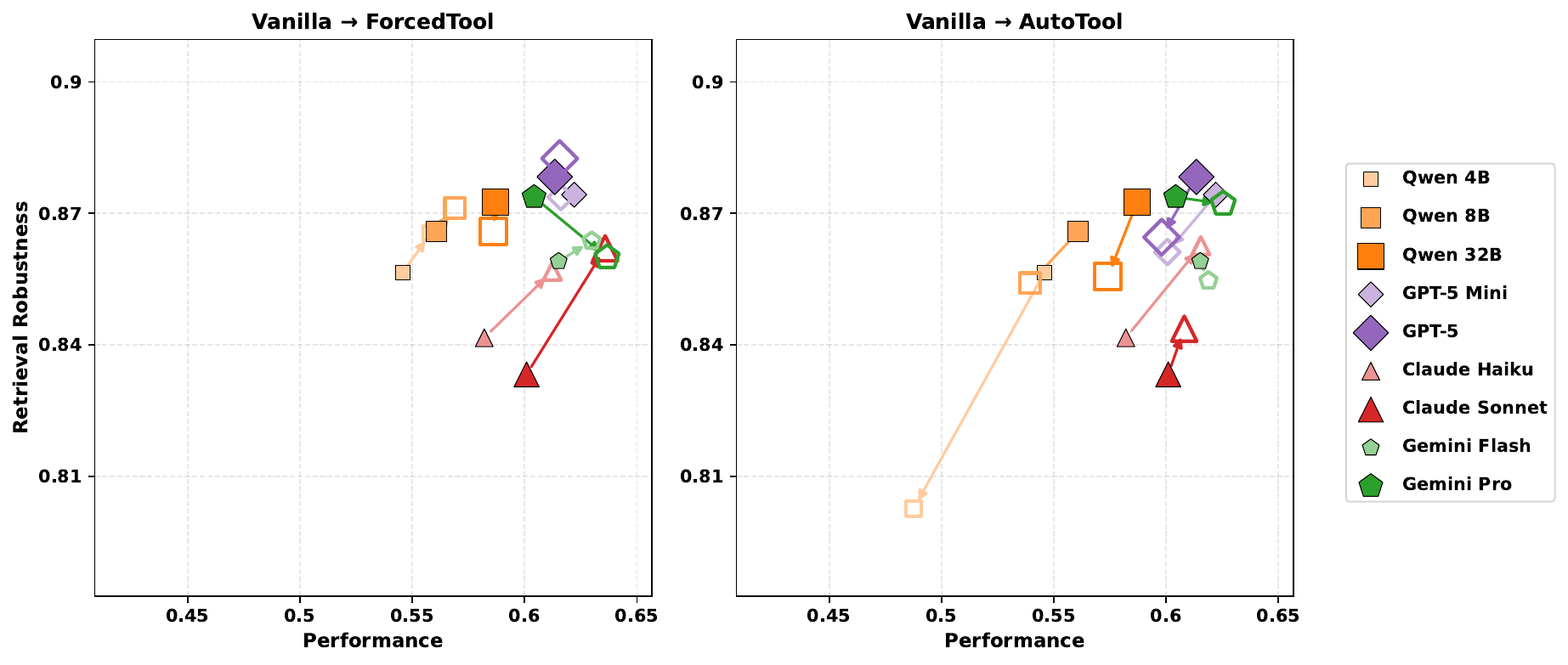}
    \caption{Effects of tool usage. Models are required to call the given retrieval tool in ForcedTool setting and can decide whether to call the retrieval tool in AutoTool setting. Retrieved documents are then provided as tool responses.}
    \label{fig:effect_tools}
\end{figure*}

Claude models consistently demonstrate better robustness and performance when retrieved documents are treated as tool responses, as shown in Figure~\ref{fig:effect_tools}.
This suggests that Claude models are highly optimized for tool-based interactions, which is further evidenced by their strong performance in agentic applications like Claude Code~\citep{claudecode2025}.
Consequently, we hypothesize that injecting additional context through tool calls and responses, even through dummy tools, is likely a more effective integration strategy for Claude.

In contrast, Qwen and GPT models exhibit noticeable drops in robustness and performance when tool calls are not strictly required.
Compared to Claude and Gemini models, Qwen and GPT models utilize the retrieval tool significantly less frequently.
For instance, in the same setup, while Claude and Gemini models trigger retrieval over 99\% of the time, Qwen-4B and GPT-5 do so only 87\% and 33\% of the time, respectively.
Notably, this lower frequency of tool invocation does not inherently account for the drop of robustness. 
When the retrieval tool is bypassed, these models rely on their internal knowledge and should, in theory, produce outputs similar to their non-RAG baselines.
We instead posit that the reduced robustness of Qwen and GPT under the AutoTool setting stems from a lack of robustness to the additional tool schema in the context, leading to varying outputs.

\subsection{Oracle Performance}

\begin{wrapfigure}[16]{r}{0.5\textwidth}
    \vspace{-2em}
    \centering
    \includegraphics[width=0.45\textwidth]{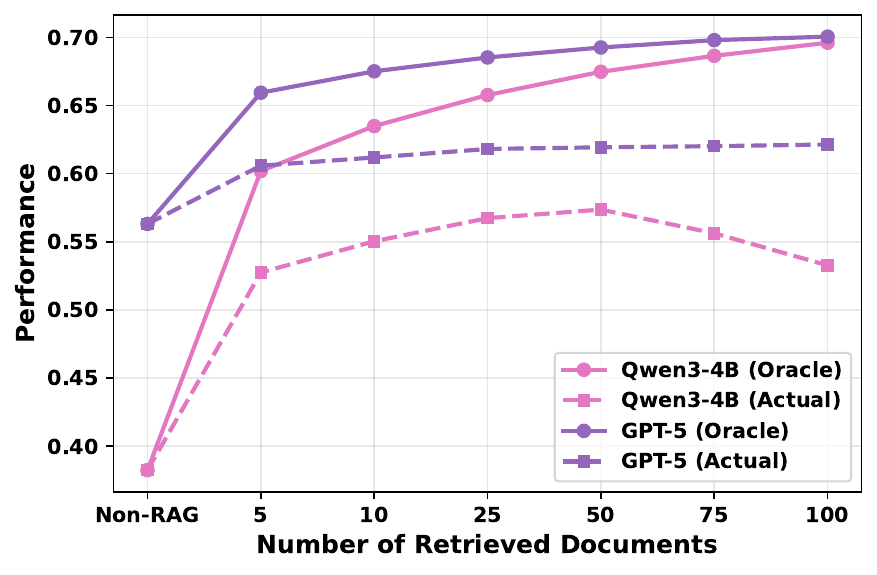}
    \caption{Average task performance of models using vanilla prompting with actual retrieval robustness (Actual) and perfect retrieval robustness (Oracle) across $k$s.
    Perfect retrieval robustness means the model would not produce outputs that are worse than it has previously produced with smaller values of $k$.}
    \label{fig:oracle_performance}
\end{wrapfigure}
Lastly, we investigate the oracle performance when models have perfect retrieval robustness across $k$s.
Concretely, for each sample, we keep the best output the model generates so far until the current $k$ across all retrieval orders.

As illustrated by Figure~\ref{fig:oracle_performance}, having perfect retrieval robustness can lead to 8\% and 15\% of absolute performance improvement for GPT-5 and Qwen-4B.
Notably, Qwen-4B can even reach similar performance to GPT-5, suggesting the importance of retrieval robustness.

\subsection{Performance-Weighted Robustness}

\begin{figure}[t]
    \centering
    \includegraphics[width=0.96\textwidth]{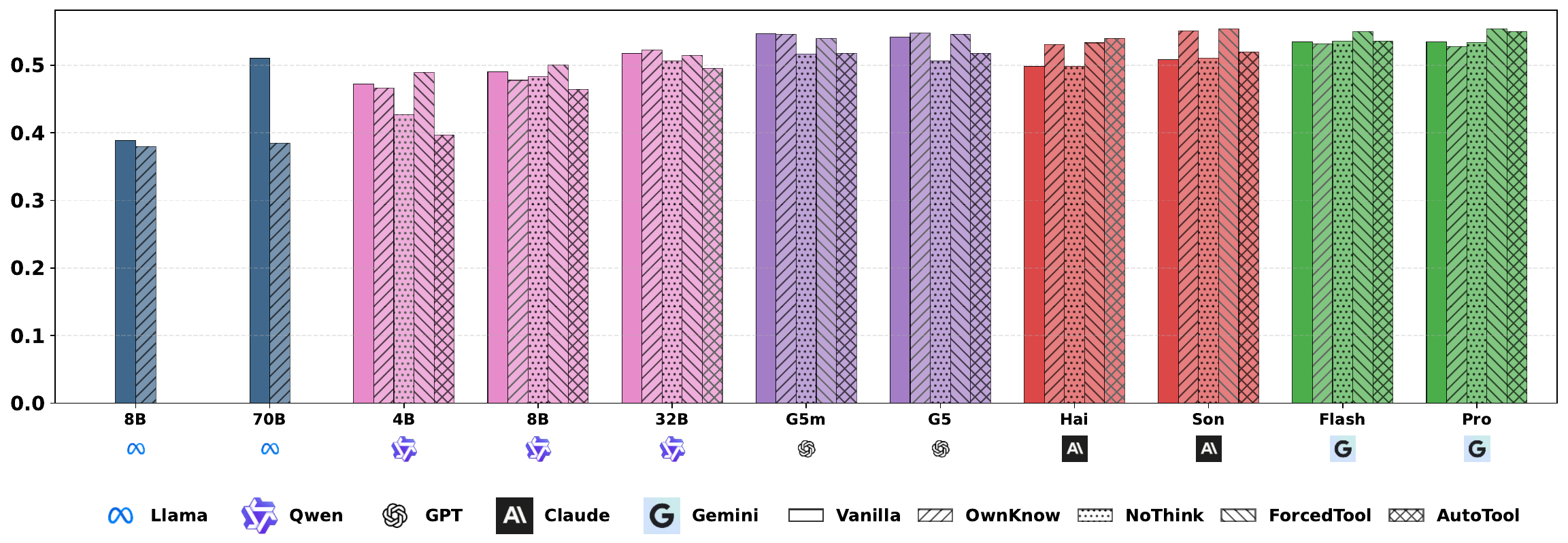}
    \caption{Performance-Weighted Robustness (PWR) across all five prompting setups. PWR combines robustness and task performance into a single score, penalizing models that achieve high robustness simply by being consistently wrong. Llama models are only tested with vanilla and OwnKnow setups.}
    \label{fig:pwr_all_setups}
\end{figure}

While retrieval robustness is orthogonal to task performance and provides complementary dimensions for model evaluation, a model that consistently gives incorrect answers can achieve perfect robustness.
To mitigate confusion for model selection, we consider \textbf{Performance-Weighted Robustness (PWR)}, which jointly accounts for both retrieval robustness and task performance:
\begin{equation}
    \text{PWR} = (\text{NDR} \cdot \text{RSR} \cdot \text{ROR})^{1/3} \times \text{Performance}
\end{equation}
where Performance is the average task score across all retrieval sizes, orders, retrievers, and datasets.
Multiplying retrieval robustness by performance ensures that only models that are both robust and accurate receive high scores.

Figure~\ref{fig:pwr_all_setups} shows PWR across all five prompting setups.
Llama-3.3-70B, under the OwnKnow setup, is significantly penalized for its low task performance despite maintaining good robustness compared to the vanilla setup (Figure~\ref{fig:effect_ownknow_nothink}).
While GPT-5 has almost the same task performance under vanilla and OwnKnow setups, the better retrieval robustness leads to higher PWR.
Among the setups, \textbf{ForcedTool yields the highest PWR} for most models, further highlighting the benefit of formulating retrieval as tool calls and tool responses.

\section{Conclusions}

We introduce retrieval robustness metrics---no-degradation rate, retrieval size robustness, and retrieval order robustness---to quantify how reliably LLMs handle queries via RAG.
A realistic benchmark of 1{,}891 samples is compiled, spanning five datasets across three task categories, with documents retrieved using both sparse and dense retrievers.
Our experiments with 11 LLMs from 5 families reveal that models generally achieve high retrieval robustness, but robustness varies substantially across tasks, with long-form generation being notably less robust than short-form QA and fact verification.
Through four additional prompting strategies, we find that reasoning-trained models such as Qwen and GPT suffer notable robustness declines when reasoning is disabled, even on simple tasks, and that tool-based retrieval integration benefits Claude models but hurts Qwen and GPT.
We hope our benchmark and findings inspire further research on robust RAG systems.

\section*{Reproducibility Statement}

Questions in our benchmark come from Natural Questions~\citep{kwiatkowski-etal-2019-natural} (Hugging Face\footnote{\url{https://huggingface.co/datasets/google-research-datasets/natural_questions}}), HotpotQA~\citep{yang-etal-2018-hotpotqa} (Hugging Face\footnote{\url{https://huggingface.co/datasets/hotpotqa/hotpot_qa}}), MS MARCO~\citep{bajaj2018msmarco} (Hugging Face\footnote{\url{https://huggingface.co/datasets/microsoft/ms_marco}}), SciFact-Open~\citep{wadden-etal-2022-scifact} (Hugging Face\footnote{\url{https://huggingface.co/datasets/davidheineman/scifact-open}}), and WixQA~\citep{wixqa} (Hugging Face\footnote{\url{https://huggingface.co/datasets/Wix/WixQA}}).
We will release code to reproduce our benchmark, including the script for obtaining sampled questions and retrieval corpus.
We will also release the script for calculating retrieval robustness metrics given model outputs.

Most of our experiment settings can be found in Section~\ref{sec:benchmark}, with inference details available in Appendix~\ref{app:inference}.
Furthermore, we include the prompt templates we use in our experiment in Appendix~\ref{app:prompt}.


\bibliography{colm2026_conference}
\bibliographystyle{colm2026_conference}

\appendix
\section{Additional Analysis}

\subsection{Soft Trend-based Retrieval Size Robustness}
\label{app:trend_rsr}

We consider a softer, trend-based alternative to retrieval size robustness, which measures the overall monotonic trend of the performance trajectory as retrieval size increases. 
Formally, for a fixed example $q$, document order $o$, and the set of retrieval sizes $K = \{k_1, \dots, k_n\}$ with $k_1 < \cdots < k_n$, we define:
\begin{equation}
    \text{Trend-RSR}_{(q,o)} = \frac{\tau_{(q,o)} + 1}{2}
\end{equation}
where $\tau_{(q,o)}$ is Kendall's rank correlation coefficient between the retrieval-size index and the corresponding performance scores:
\begin{equation}
    \tau_{(q,o)} = \frac{1}{\binom{n}{2}} \sum_{\substack{i < j}} \text{sgn}\bigl(f(q, k_j, o) - f(q, k_i, o)\bigr)
\end{equation}
The $(\tau + 1)/2$ rescaling maps the correlation from $[-1, 1]$ to $[0, 1]$: a score of 1.0 indicates a perfectly non-decreasing trajectory (performance never drops as $k$ increases), 0.5 indicates no systematic trend, and values below 0.5 indicate a predominantly decreasing trend.

\begin{figure}[h]
    \centering
    \includegraphics[width=0.96\textwidth]{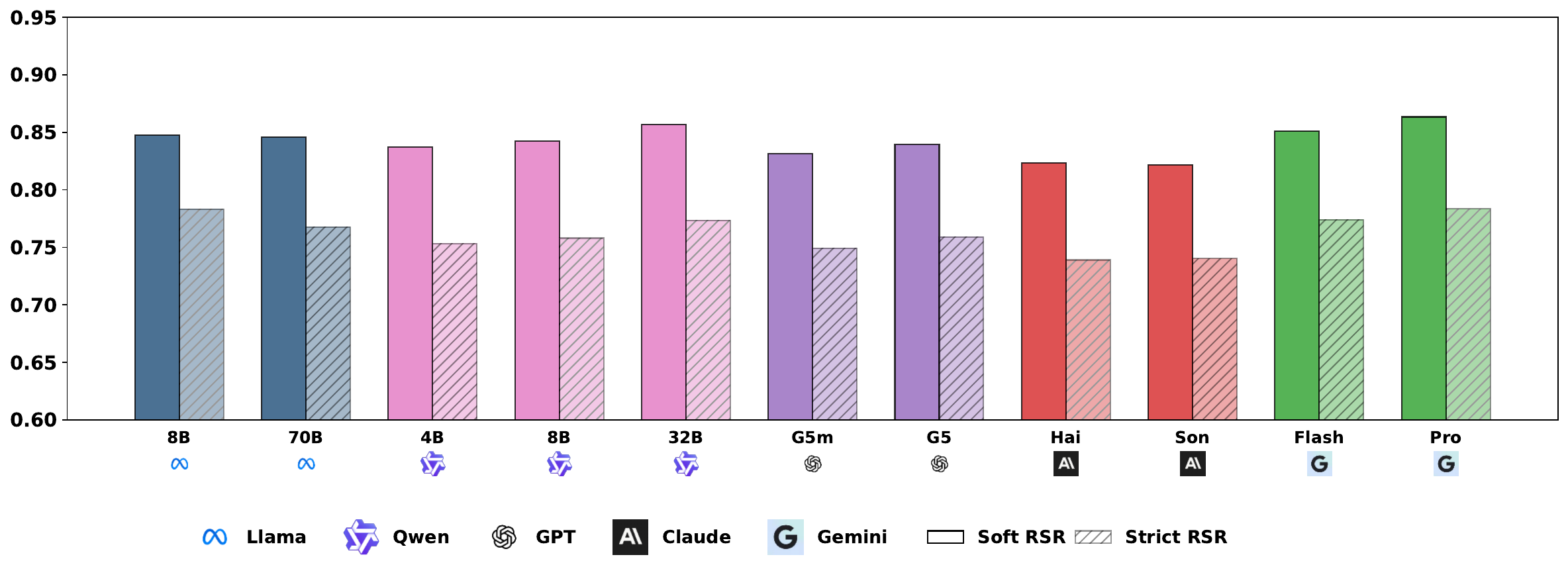}
    \caption{Soft RSR vs.\ Strict RSR per model. Models have higher RSR under the softer formulation, while their ranks are consistent across the two formulations.}
    \label{fig:trend_rsr}
\end{figure}

\paragraph{Results.}
Figure~\ref{fig:trend_rsr} compares soft RSR against the original strict RSR for the vanilla prompting setup.
Soft RSR is consistently higher than strict RSR, as expected since it tolerates local dips that the strict formulation penalizes.
Notably, the \emph{model ranking is preserved}: models that score high (or low) on strict RSR also score high (or low) on Soft RSR.

\subsection{WixQA with LLM Judge Evaluation}
\label{app:wixqa_llm_judge}

We use ROUGE-2 recall instead of LLM judges for WixQA in the main paper due to the high cost of running LLM judges on the full dataset.
Here we conduct a supplementary analysis using LLM-based fact recall for evaluation on a subset of WixQA.

\paragraph{Setup.}

We decompose each reference answer into atomic facts using Claude Sonnet 4.5, then use three models (GPT-5-mini, Claude Haiku 4.5, and Gemini 2.5 Flash) to determine whether each reference fact is covered by the model's prediction.
The per-example fact recall score is the fraction of reference facts judged as covered, averaged across the three judges.
We evaluate the first 50 examples across all 11 models, all 6 retrieval sizes, all 3 document orders, using the BM25 retriever under the vanilla prompting setup.

\paragraph{Robustness Comparison.}

\begin{figure}[h]
    \centering
    \includegraphics[width=0.96\textwidth]{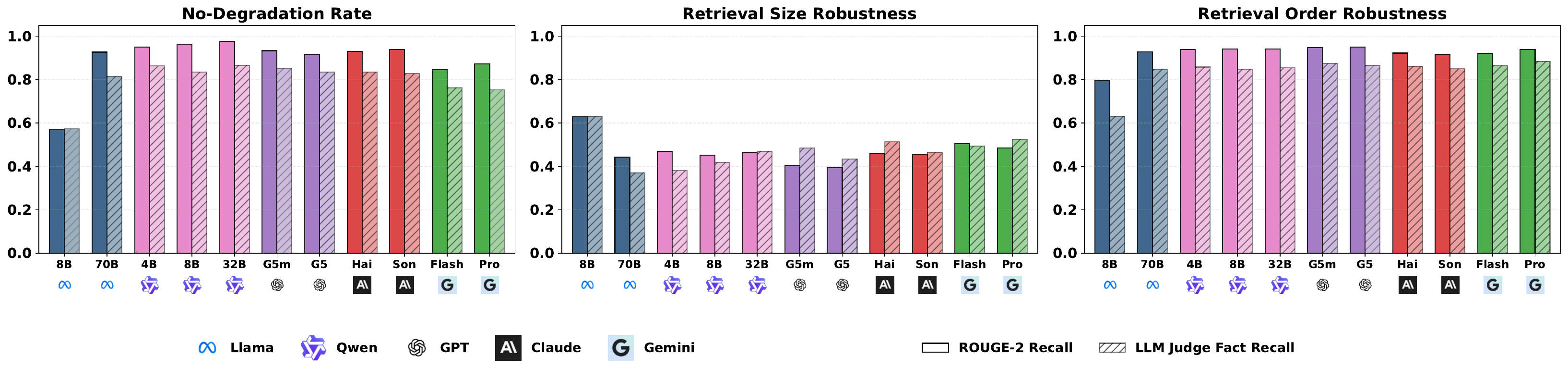}
    \caption{Robustness metrics on WixQA (50-example subset, BM25, vanilla setup) computed using ROUGE-2 recall (solid) vs.\ LLM-judge fact recall averaged over 3 judges (hatched). Low RSR on long-form QA is observed across both task evaluation metrics.}
    \label{fig:wixqa_rouge_vs_judge}
\end{figure}

Figure~\ref{fig:wixqa_rouge_vs_judge} compares the three robustness metrics computed using ROUGE-2 recall versus LLM-judge fact recall on the subset.
The relative model rankings are moderately preserved across both metrics (Spearman $\rho = 0.79$, $0.59$, and $0.45$ for the three robustness metrics).

Overall, long-form QA remains the task where models face the most substantial retrieval robustness challenges, with RSR consistently around 0.47 regardless of the scoring metric.

\subsection{Agreement Level among LLM Judges}
\label{app:judge_agreement}

For open-domain QA evaluation, we use three LLM judges from different model families (GPT-5-mini, Claude Haiku 4.5, and Gemini 2.5 Flash) to determine answer correctness via majority vote (\S\ref{sec:data_and_eval}).
We investigate if the three judges produce consistent judgment, as a proxy measure of LLM judge quality.
Across all models, datasets, retrieval sizes, document orders, and retrievers in our benchmark, the judges collectively annotate 610{,}500 samples.

As shown in Table~\ref{tab:judge_agreement}, the three judges achieve high inter-annotator agreement: Fleiss' $\kappa = 0.83$ and an average pairwise agreement rate of 91.8\%.

\begin{table}[h]
\centering
\small
\caption{Inter-judge agreement statistics on open-domain QA evaluation.}
\label{tab:judge_agreement}
\begin{tabular}{llcc}
\toprule
\textbf{Judge 1} & \textbf{Judge 2} & \textbf{Agreement} & \textbf{Cohen's $\kappa$} \\
\midrule
GPT-5-mini & Claude Haiku 4.5 & 0.906 & 0.807 \\
GPT-5-mini & Gemini 2.5 Flash & 0.946 & 0.887 \\
Claude Haiku 4.5 & Gemini 2.5 Flash & 0.901 & 0.797 \\
\midrule
\multicolumn{2}{l}{\textbf{Overall (all 3 judges)}} & & \\
\multicolumn{2}{l}{\quad Fleiss' $\kappa$} & \multicolumn{2}{c}{0.830} \\
\multicolumn{2}{l}{\quad Avg pairwise agreement} & \multicolumn{2}{c}{0.918} \\
\bottomrule
\end{tabular}
\end{table}

\section{Dataset Breakdown Results}

\subsection{Retrieval Robustness}

\begin{figure}
    \centering
    \includegraphics[width=0.96\textwidth]{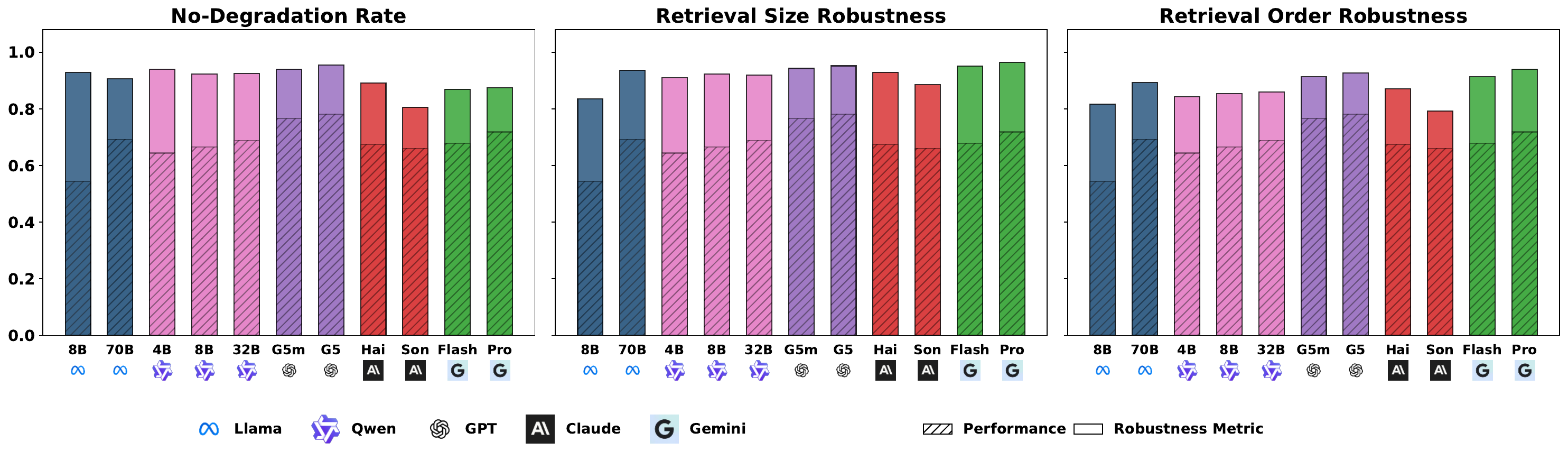}
    \caption{The three retrieval robustness metrics and task performance of experimented LLMs using vanilla prompting on Natural Questions.}
    \label{fig:robustness_separate_result_natural_questions}
\end{figure}

\begin{figure}
    \centering
    \includegraphics[width=0.96\textwidth]{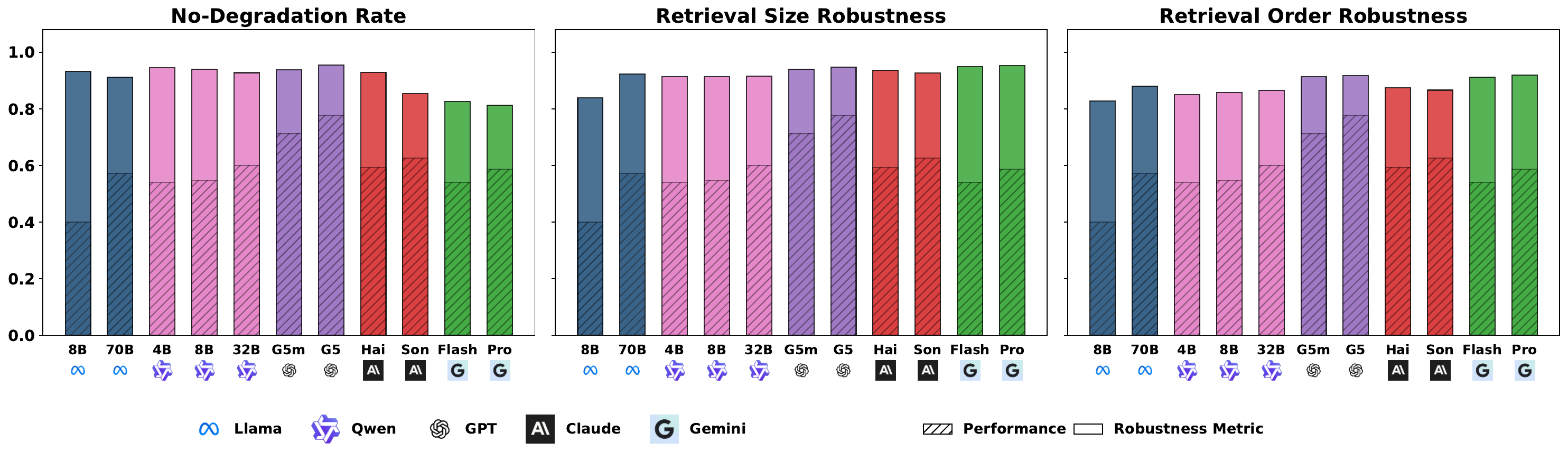}
    \caption{The three retrieval robustness metrics and task performance of experimented LLMs using vanilla prompting on HotpotQA.}
    \label{fig:robustness_separate_result_hotpot_qa}
\end{figure}

\begin{figure}
    \centering
    \includegraphics[width=0.96\textwidth]{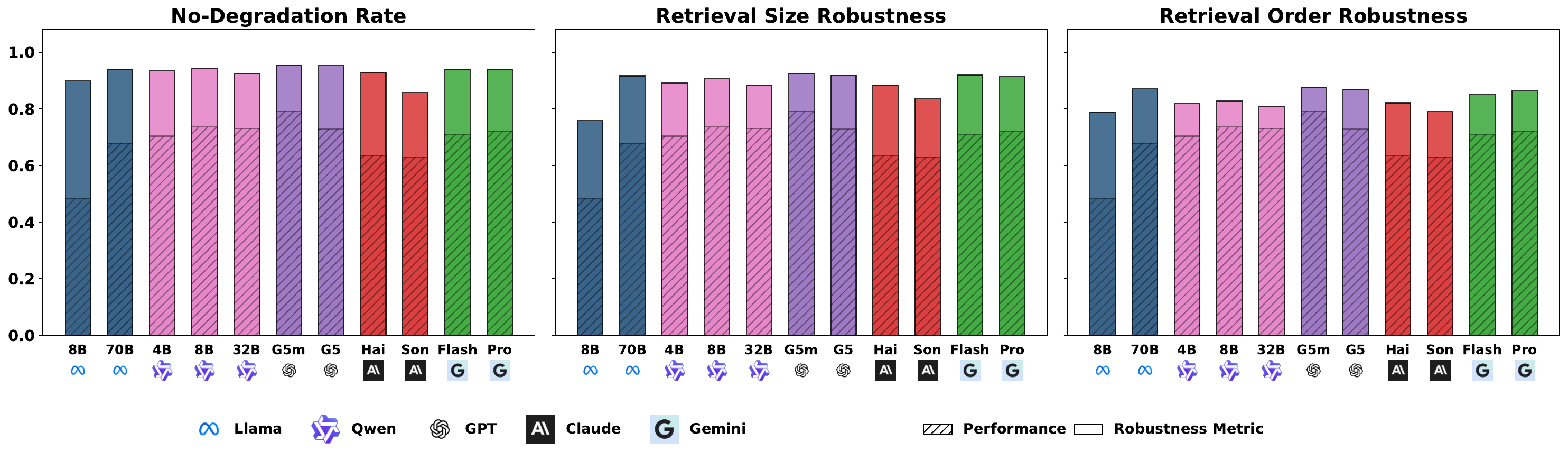}
    \caption{The three retrieval robustness metrics and task performance of experimented LLMs using vanilla prompting on MS MARCO.}
    \label{fig:robustness_separate_result_msmarco}
\end{figure}

\begin{figure}
    \centering
    \includegraphics[width=0.96\textwidth]{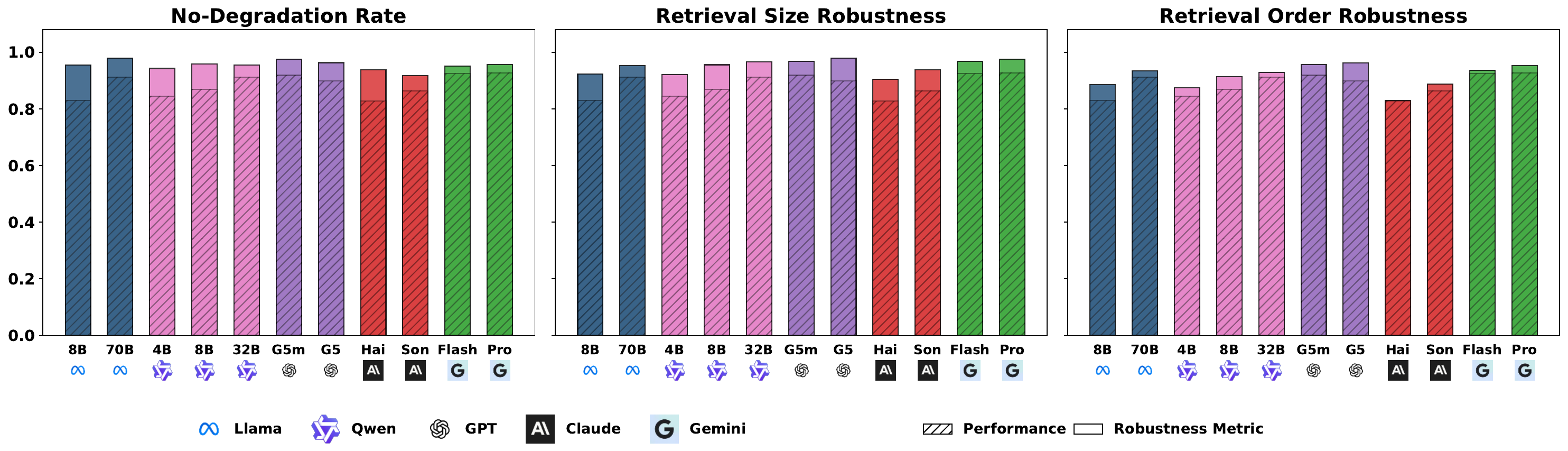}
    \caption{The three retrieval robustness metrics and task performance of experimented LLMs using vanilla prompting on SciFact-Open.}
    \label{fig:robustness_separate_result_scifact}
\end{figure}

\begin{figure}
    \centering
    \includegraphics[width=0.96\textwidth]{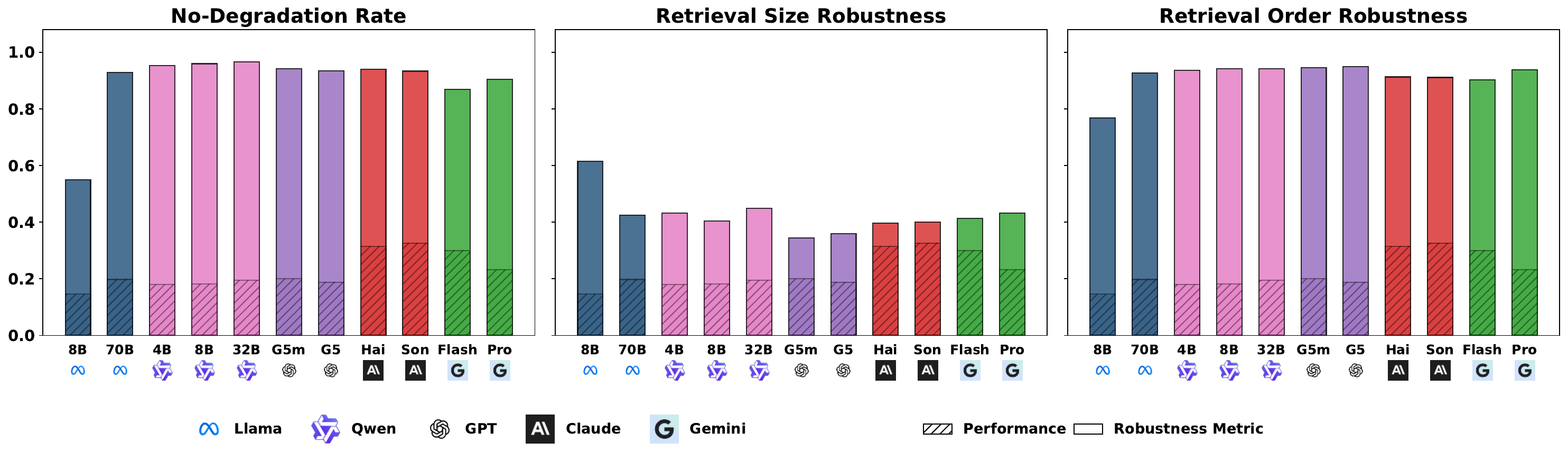}
    \caption{The three retrieval robustness metrics and task performance of experimented LLMs using vanilla prompting on WixQA.}
    \label{fig:robustness_separate_result_wixqa}
\end{figure}

We show the retrieval robustness metrics and average RAG performance in Figures~\ref{fig:robustness_separate_result_natural_questions}, \ref{fig:robustness_separate_result_hotpot_qa}, \ref{fig:robustness_separate_result_msmarco}, \ref{fig:robustness_separate_result_scifact}, and \ref{fig:robustness_separate_result_wixqa}.

\subsection{Prompting Strategy Effects}
\label{app:effect_per_dataset}

\begin{figure}
    \centering
    \includegraphics[width=0.96\textwidth]{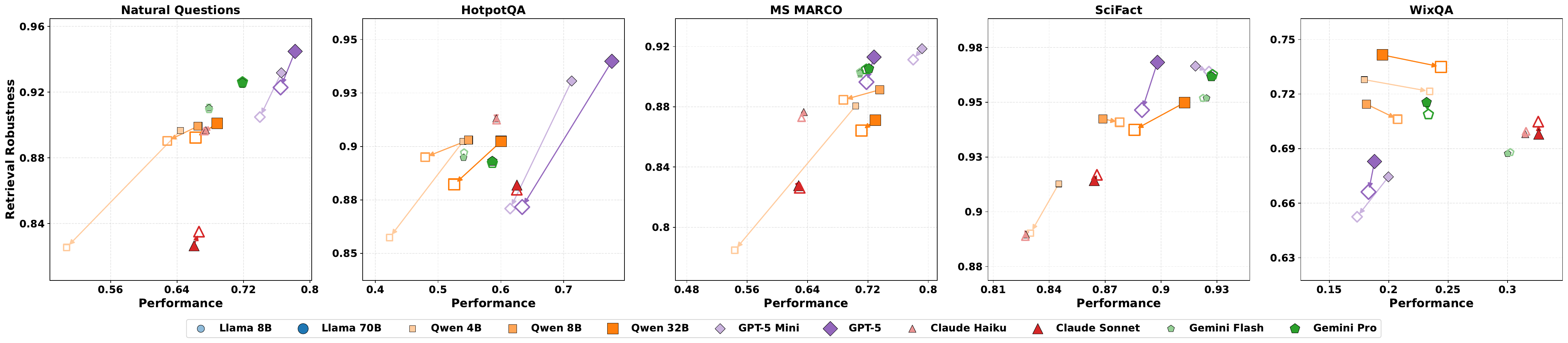}
    \caption{Effects of \textbf{NoThink} on different datasets.}
    \label{fig:effect_nothink_per_dataset}
\end{figure}

\begin{figure}
    \centering
    \includegraphics[width=0.96\textwidth]{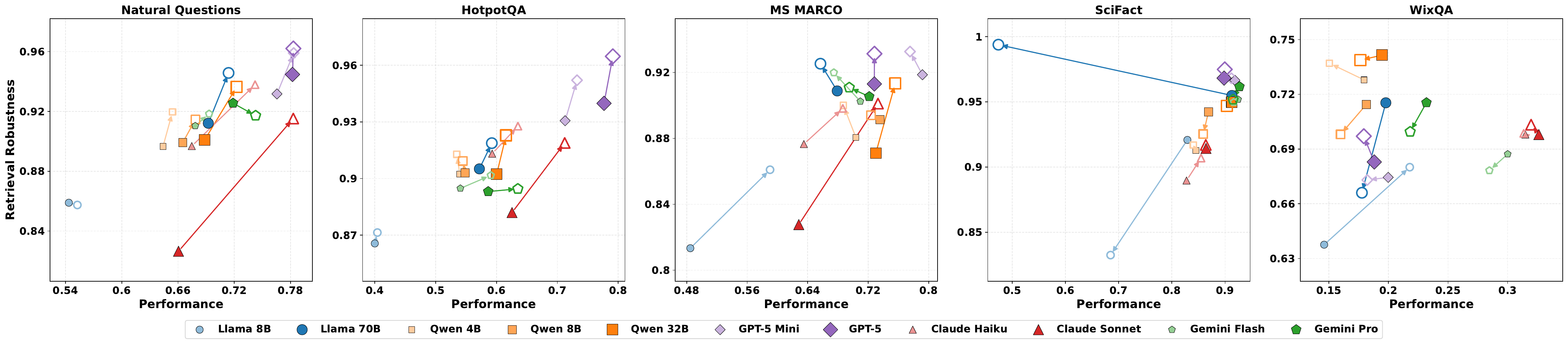}
    \caption{Effects of \textbf{OwnKnow} on different datasets.}
    \label{fig:effect_ownknow_per_dataset}
\end{figure}

\begin{figure}
    \centering
    \includegraphics[width=0.96\textwidth]{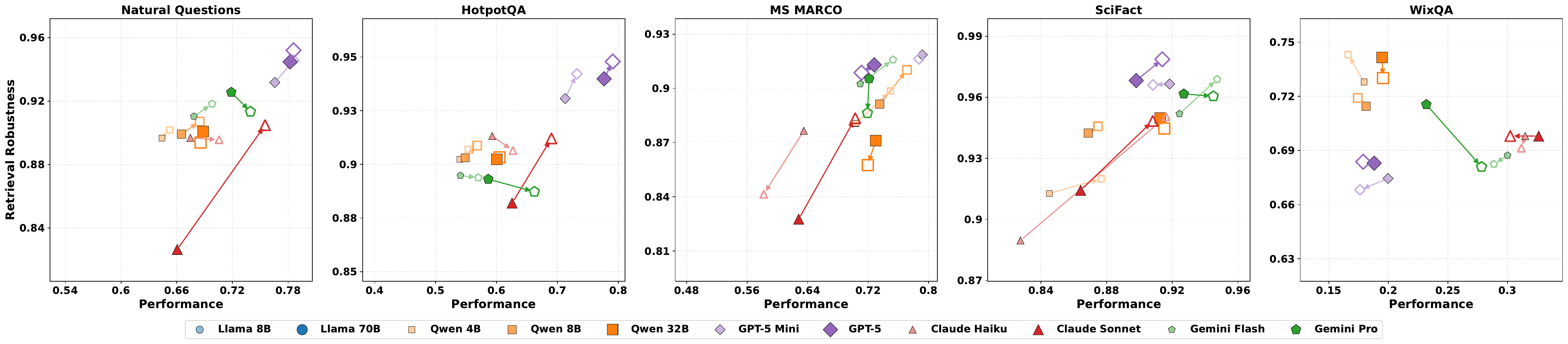}
    \caption{Effects of \textbf{ForcedTool} on different datasets.}
    \label{fig:effect_forcedtool_per_dataset}
\end{figure}

\begin{figure}
    \centering
    \includegraphics[width=0.96\textwidth]{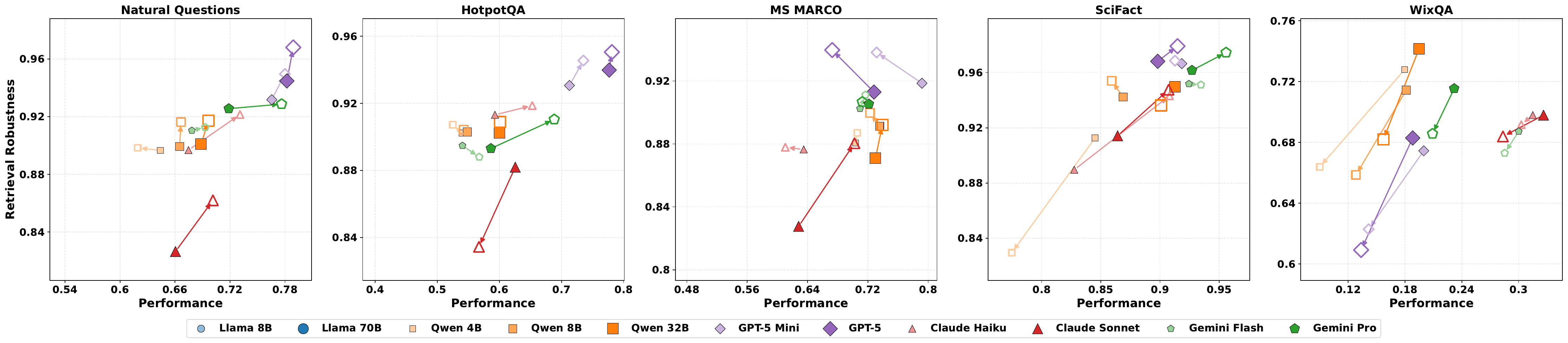}
    \caption{Effects of \textbf{AutoTool} on different datasets.}
    \label{fig:effect_autotool_per_dataset}
\end{figure}

We show the effects of \textbf{NoThink}, \textbf{OwnKnow}, \textbf{ForcedTool}, and \textbf{AutoTool} in Figures~\ref{fig:effect_nothink_per_dataset}, \ref{fig:effect_ownknow_per_dataset}, \ref{fig:effect_forcedtool_per_dataset}, and \ref{fig:effect_autotool_per_dataset}.

\subsection{Performance-Weighted Robustness}

\begin{figure}
    \centering
    \includegraphics[width=0.96\textwidth]{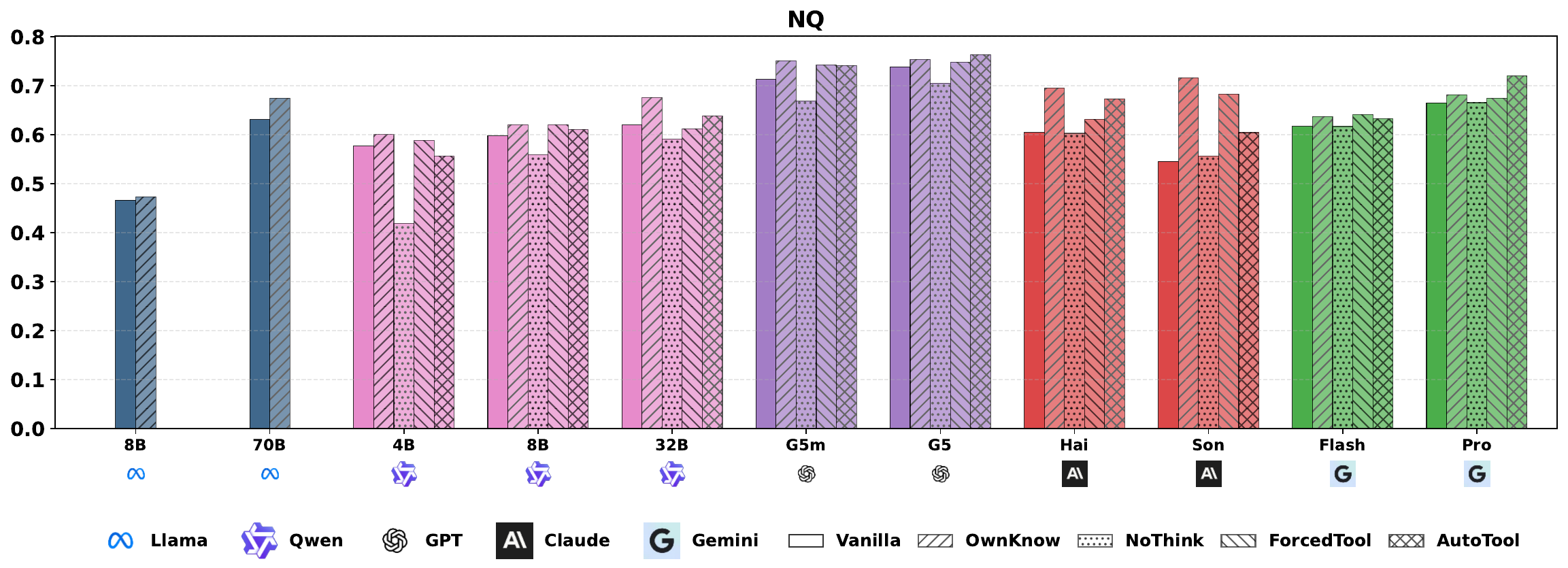}
    \caption{Performance-weighted robustness on Natural Questions.}
    \label{fig:pwr_natural_questions}
\end{figure}

\begin{figure}
    \centering
    \includegraphics[width=0.96\textwidth]{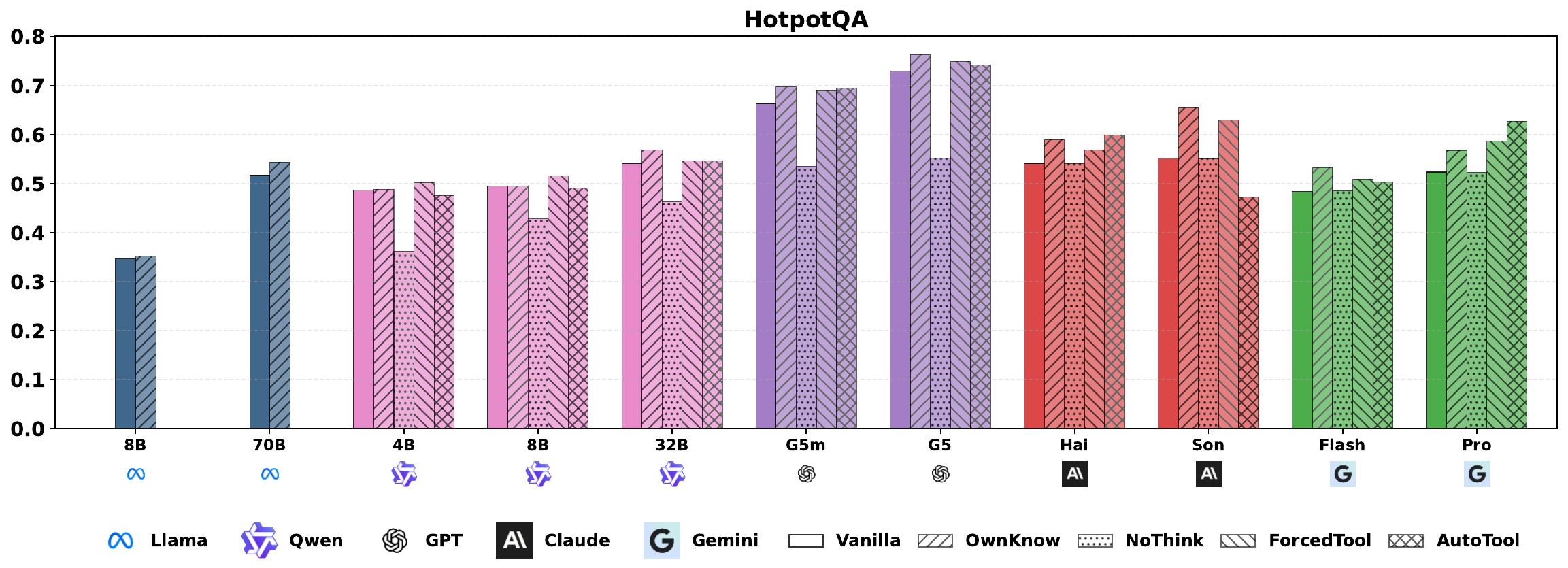}
    \caption{Performance-weighted robustness on HotpotQA.}
    \label{fig:pwr_hotpot_qa}
\end{figure}

\begin{figure}
    \centering
    \includegraphics[width=0.96\textwidth]{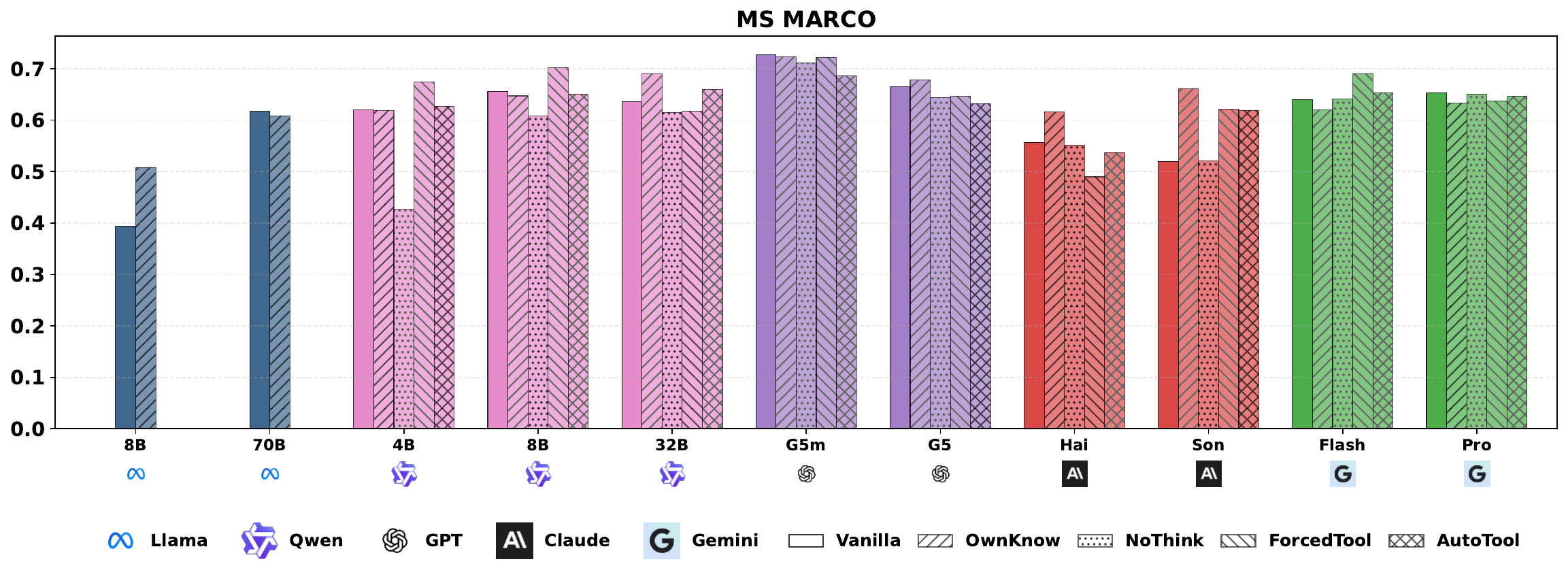}
    \caption{Performance-weighted robustness on MS MARCO.}
    \label{fig:pwr_msmarco}
\end{figure}

\begin{figure}
    \centering
    \includegraphics[width=0.96\textwidth]{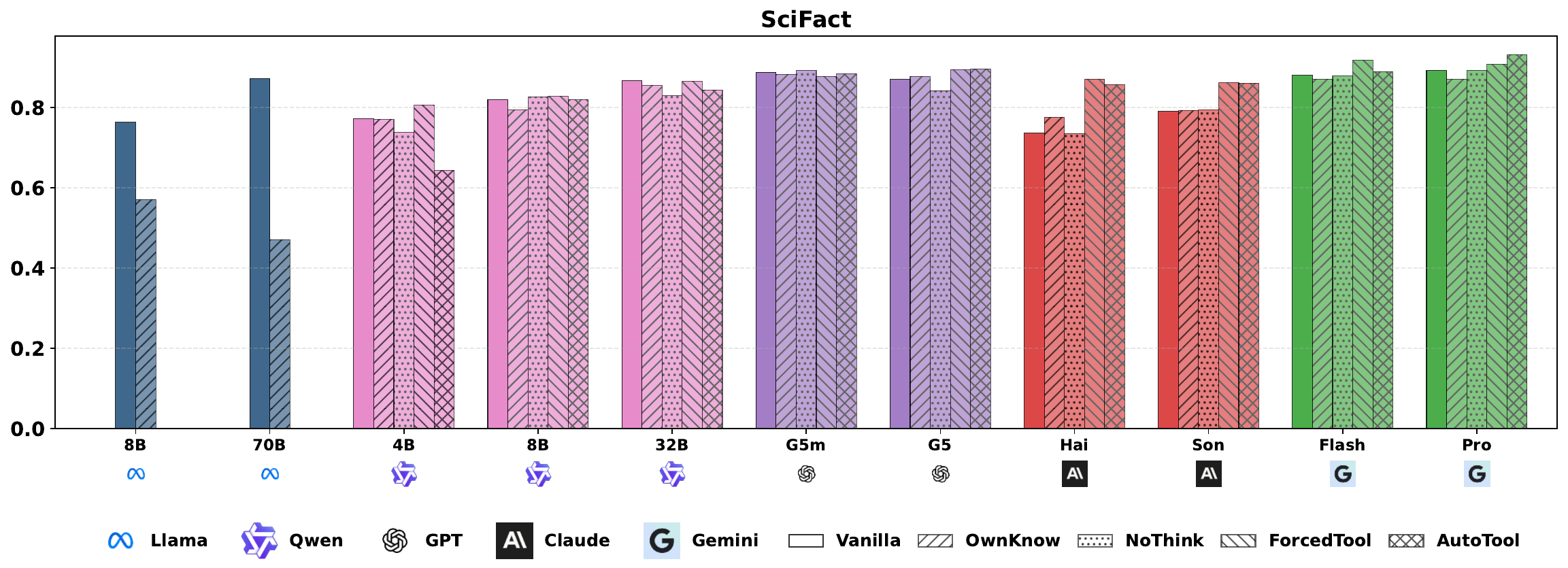}
    \caption{Performance-weighted robustness on SciFact-Open.}
    \label{fig:pwr_scifact}
\end{figure}

\begin{figure}
    \centering
    \includegraphics[width=0.96\textwidth]{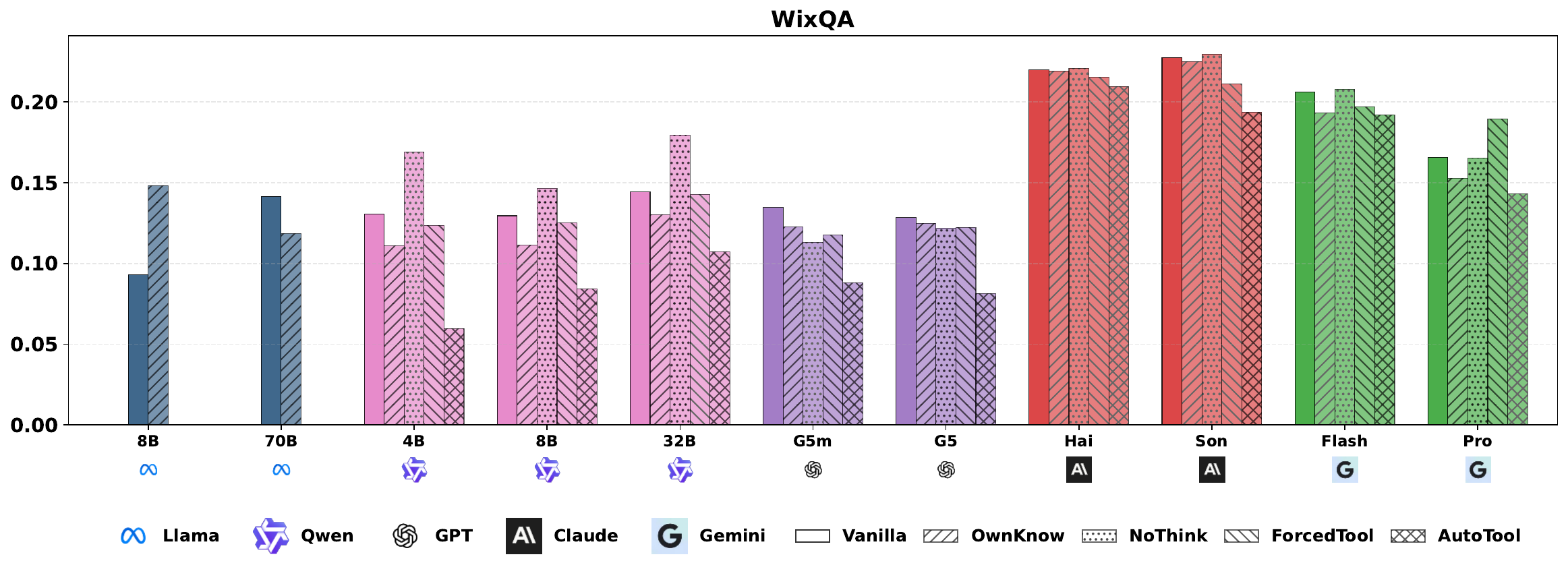}
    \caption{Performance-weighted robustness on WixQA.}
    \label{fig:pwr_wixqa}
\end{figure}

We show the performance-weighted robustness on the 5 datasets in Figures~\ref{fig:pwr_natural_questions}, \ref{fig:pwr_hotpot_qa}, \ref{fig:pwr_msmarco}, \ref{fig:pwr_scifact}, and \ref{fig:pwr_wixqa}.

\section{Inference Setup}
\label{app:inference}

\paragraph{Inference Parameters.}

Due to the computational cost and running time, we use greedy decoding and perform inference with each model under each setup once.
However, we note that for proprietary models, setting temperature to 0 would not result in deterministic decoding.
During inference, models are allowed to generate at most 4096 tokens on all datasets, except for WixQA where we allow models to generate 8192 tokens.

\paragraph{Inference Infrastructure.}

For models with publicly available weights,
we use vLLM for more efficient inference~\citep{kwon2023efficient}, and our experiments are conducted on compute nodes with 8 H100 GPUs.

\section{The Use of LLMs}

We use LLMs (Gemini, Claude) to polish writing, including correcting grammar errors, making language more concise, and plotting figures.
\section{Prompt Templates}
\label{app:prompt}

The prompt templates (in jinja2 format) used in our experiments can be found at the end of the Appendix.

\onecolumn
\subsection*{\texttt{non\_rag\_qa.j2}}
\begin{lstlisting}
Answer the user given question in a concise manner. Indicate the answer with "Answer:" and only include the answer words or phrases.<eosys>{{ question }}
\end{lstlisting}

\subsection*{\texttt{rag\_qa.j2}}
\begin{lstlisting}
Answer the user given question in a concise manner. Indicate the answer with "Answer:" and only include the answer words or phrases. The user would also provide retrieved documents that might be useful for answering the question.<eosys>Retrieved Documents:
{% for source in sources %}Document {{ loop.index }}:
{{ source }}

{% endfor %}{{ question }}
\end{lstlisting}

\subsection*{\texttt{rag\_qa\_ownknow.j2}}
\begin{lstlisting}
Answer the user given question in a concise manner. Indicate the answer with "Answer:" and only include the answer words or phrases. The user would also provide retrieved documents that might be useful for answering the question. In addition, the user would also provide your previously made answer (if available) which you can revise.<eosys>Retrieved Documents:
{% for source in sources %}Document {{ loop.index }}:
{{ source }}

{% endfor %}{{ question }}
Previous Answer: {{ open_qa_output }}
\end{lstlisting}

\subsection*{\texttt{rag\_qa\_tool.j2}}
\begin{lstlisting}

{{ SYSTEM_START }}
Answer the user given question in a concise manner. Indicate the answer with "Answer:" and only include the answer words or phrases. Access to a retriever is provided to gather relevant documents. You are allowed to call the retriever at most once.
{{ SYSTEM_END }}
{{ USER_START }}
{{ question }}
{{ USER_END }}
\end{lstlisting}

\subsection*{\texttt{answer\_match.j2}}
\begin{lstlisting}
{{ SYSTEM_START }}
You will be given a question, a list of gold answers to this question, and a predicted answer. Any answer from the gold answer list is considered a correct answer to the question. Your task is to judge whether the predicted answer can match any of the gold answers.
Note that predicted answer does not have to exactly match the gold answers. It is considered a match as long as the predicted answer is semantically equivalent to any of the gold answers.

Your response should be in the following JSON format:
{
  "rationale": "Your rationale for the judgment as to whether prediction is matching any of the gold answers.",
  "matching": 1 if the prediction matches any of the gold answers, 0 otherwise.
}
{{ SYSTEM_END }}
{{ USER_START }}
QUESTION:
{{rendered_query}}

GOLD ANSWER:
{{gold_answer}}

PREDICTION:
{{prediction}}
{{ USER_END }}
\end{lstlisting}

\twocolumn

\end{document}